\documentclass[letterpaper, 10 pt, conference]{ieeeconf}  

\usepackage{graphicx}
\usepackage{listings}
\usepackage{subfigure}
\usepackage{amsmath}
\usepackage{amsfonts}
\usepackage{bm}
\usepackage{booktabs}
\usepackage[export]{adjustbox}
\usepackage[table]{xcolor}
\usepackage{multirow}
\usepackage{makecell}                 
\usepackage[belowskip=-10pt,aboveskip=5pt]{caption}
\usepackage{algorithm}
\usepackage{algpseudocode}
\usepackage{subcaption} 
\usepackage{caption} 
\usepackage{cite}
\usepackage[numbers,sort&compress]{natbib}
\usepackage{hyperref}
\usepackage{siunitx}
\hypersetup{
    colorlinks=true, 
    linkcolor=blue, 
    citecolor=blue, 
    urlcolor=blue  
}

\IEEEoverridecommandlockouts                              

\title{\LARGE \bf
RMG: Real-Time Expressive Motion Generation with Self-collision Avoidance for 6-DOF Companion Robotic Arms
}

\author{Jiansheng LI$^{1}$, Haotian SONG$^{1}$, Jinni ZHOU$^{1*}$, Qiang NIE$^{1*}$, and Yi CAI$^{1*}$
\thanks{$^{1}$Hong Kong University of Science and Technology (Guangzhou), Guangzhou, China {\tt\small jli204@connect.hkust-gz.edu.cn}}%
\thanks{*Co-corresponding authors with equal contribution.}
}

\begin{document}

\maketitle
\thispagestyle{empty}
\pagestyle{empty}

\begin{abstract}
The six-degree-of-freedom (6-DOF) robotic arm has gained widespread application in human-coexisting environments. While previous research has predominantly focused on functional motion generation, the critical aspect of expressive motion in human-robot interaction remains largely unexplored. This paper presents a novel real-time motion generation planner that enhances interactivity by creating expressive robotic motions between arbitrary start and end states within predefined time constraints. Our approach involves three key contributions: first, we develop a mapping algorithm to construct an expressive motion dataset derived from human dance movements; second, we train motion generation models in both Cartesian and joint spaces using this dataset; third, we introduce an optimization algorithm that guarantees smooth, collision-free motion while maintaining the intended expressive style. Experimental results demonstrate the effectiveness of our method, which can generate expressive and generalized motions in under 0.5 seconds while satisfying all specified constraints.

\end{abstract}


\section{INTRODUCTION}
With the advancement of robotic intelligence, robots are increasingly deployed in human-coexisting environments. Enhancing human-robot interaction has become a growing trend, particularly in the field of companion robots \cite{bib0}. Beyond performing tasks and dialogue intelligently, expressive motions are crucial for physical interaction. Research indicates that expressive motions can convey interaction nuances beyond speech, offering subtle and engaging biological interactions that capture attention and foster acceptance  \cite{bib2}. The 6-DOF robotic arm, with its compact and highly flexible structure, is well-suited for various tasks and environmental integration, making it an excellent platform for home companion robots. Exploring how to utilize the rich postures of robotic arms to create expressive motions for better human interaction is an intriguing research topic, attracting attention from various fields such as robotics,psychology, and art\cite{bib1}.

While some research focus on explicit motion principles \cite{17}, these methods often lack generalization. Data-driven learning \cite{26} offers a solution but is hindered by the scarcity of expressive motion data for robotic arms, as most datasets are functional. Cross-modal models \cite{23} can produce high-quality motions but are typically too large for real-time use. Additionally, most methods ignore physical constraints, limiting them to simulations. Overall, current approaches fail to balance generalization, expressiveness, real-time performance, and real-world applicability \cite{bib8}. Generating expressive motion for robotic arms faces several challenges.

\begin{figure}[t]
    \centering
    \includegraphics[width=0.95\linewidth]{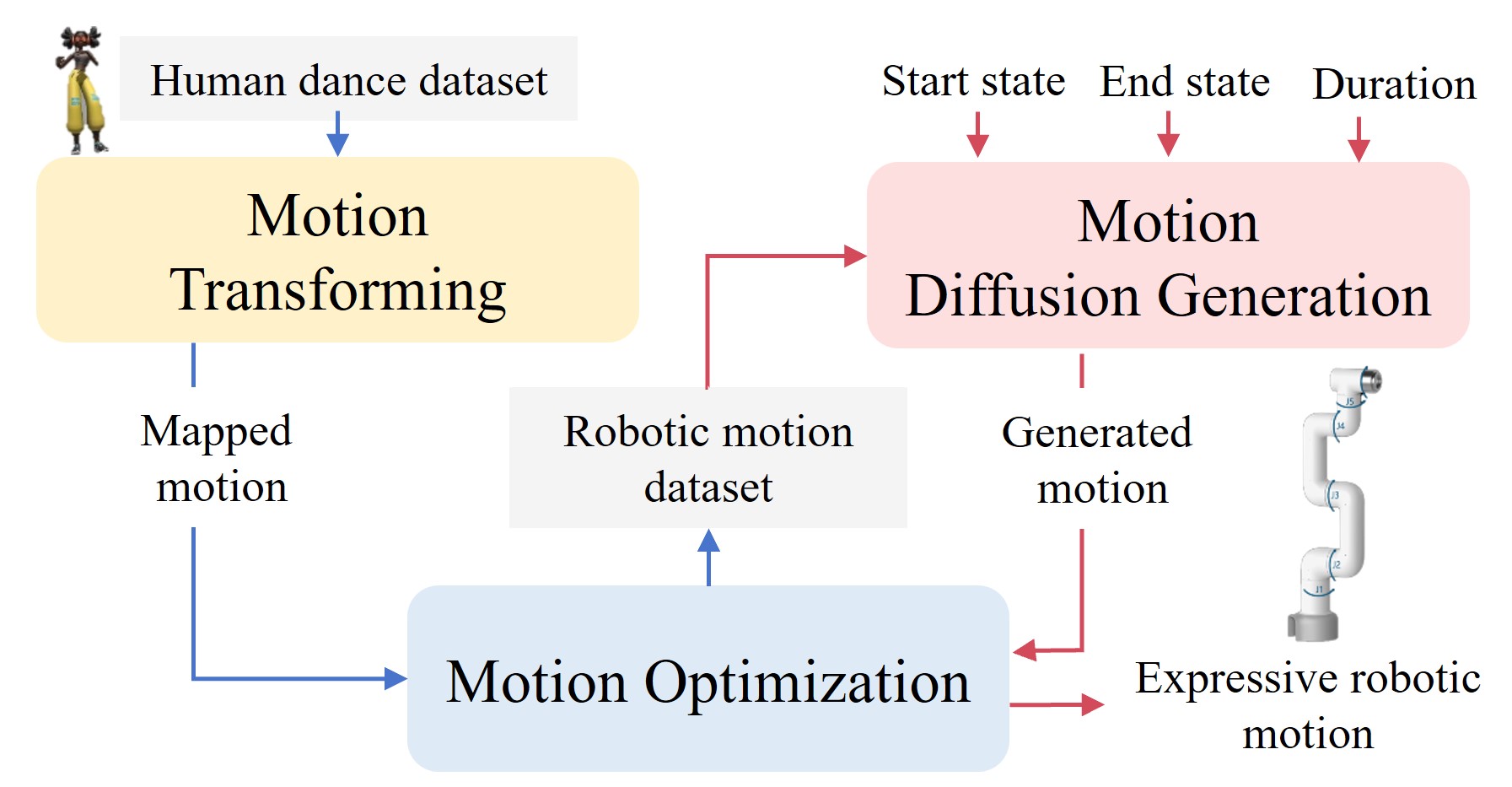}
    \caption{ \textbf{ Overview of RMG.} Blue arrows indicate the robotic dataset construction pipeline, while red arrows represent the robotic motion generation process.}
    \label{fig:1}
\end{figure}
To address these challenges, this work develops a real-time and expressive motion generation system for 6-DOF robotic arms, enabling motion planning between arbitrary start and end points within specified time constraints. This allows the robot to execute expressive motions while completing tasks. Given the lack of suitable datasets, we designed a motion transforming algorithm that creates an expressive robotic arm dataset through mapping the human dance motions. To generalize these robotic motions, we employ a diffusion model in both Joint Space and Cartesian Space, and we investigate the performance differences between the two models, highlighting the importance of global pose information. To ensure collision-free, feasible, and smooth motions, we developed an optimization algorithm that outperforms existing methods in both real-time performance and motion style preservation. Experimental results demonstrate that our method can generate expressive smooth and collision-free motions within 0.5 seconds. Finally, a prototype showcases how the system integrates with speech and sensors to enhance human-robot interaction.

In summary, our approach enables expressive, collision-free, and smooth motion for robotic arms, inspired by human dance styles. The method is applicable to diverse robotic systems equipped with arms, enhancing human-robot interaction.
\begin{itemize}  
    \item A motion transforming algorithm that maps human dance to expressive robotic motions,  creating a high-quality robotic dataset, which will be open-source.
    \item A real-time motion planner based on a diffusion model, balancing generalization and expressiveness.
    \item A real-time motion optimization algorithm that preserves the style of the motion trajectory. 
\end{itemize}


\begin{figure*}[t]
    \centering
    \includegraphics[width=0.90\linewidth]{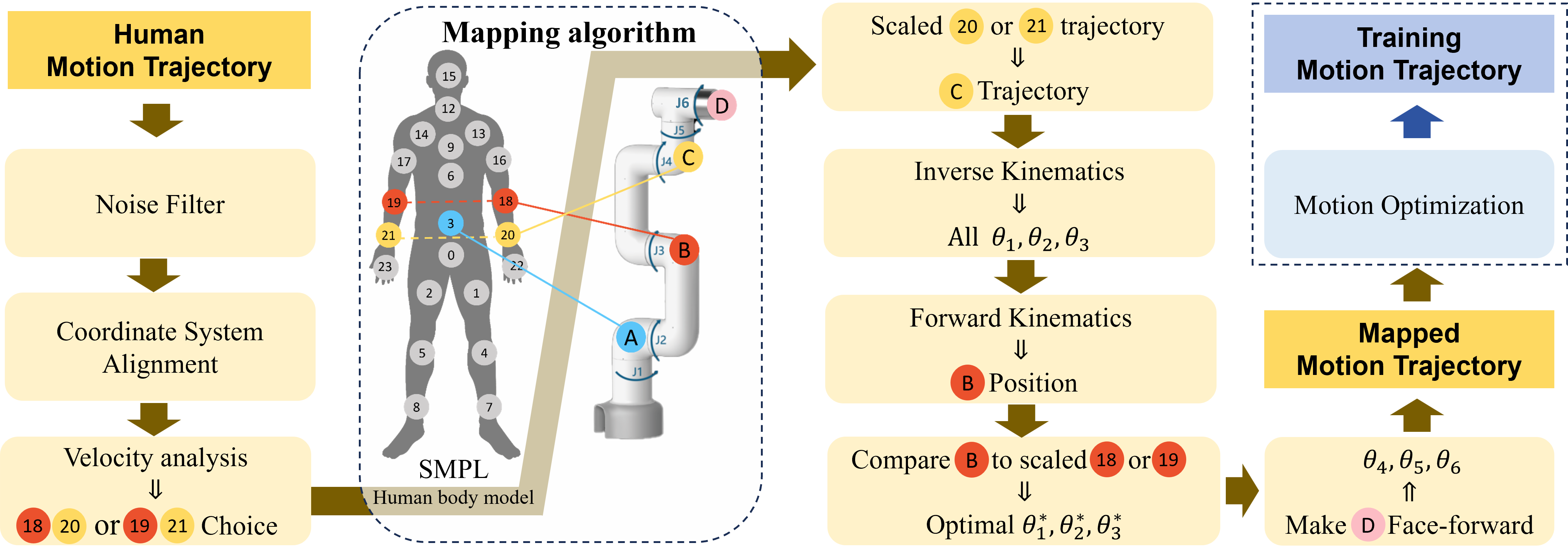}
    \caption{\textbf{Motion transforming Process.} After filtering and coordinate aligning the human motion trajectory, the total velocity at Points 20 and 21 decide the left or right trajectory. The chosen trajectory is then scaled and mapped into the robot arm's joint space. Finally, motion optimization generates a smooth, collision-free, and feasible trajectory.}
    \label{fig:2} 
\end{figure*}
\section{RELATED WORK}


\subsection{Motion Design and Mapping by Experts}
Traditional approaches to designing expressive robot motions rely on expert-driven methods. Experts use animation principles \cite{bib2}, Laban analysis \cite{15}, and biomechanics \cite{17} to generate pose-to-pose motions. While effective, these methods are labor-intensive. Another approach maps human motions to robotic motions, enabling rapid generation but requiring deep motion morphology understanding. For robotic arms, RoboGroove \cite{bib4} maps human torso angles to arm joints, but focusing solely on the torso may reduce expressiveness. Saviano et al. \cite{27} use PCA for human-robot motion mapping, which is innovative but lacks interpretability and controllability. These methods are also limited by predefined motion sets and fixed start and end points, hindering generalization.

\subsection{Learning-based Motion Generation}
Recent methods use multimodal inputs (e.g., music, sound, text) and techniques like diffusion models and transformers \cite{22,23,24,25} to generate high-quality motions. However, they are mostly limited to human or biological motion due to data availability. Additionally, their reliance on computationally intensive models restricts real-time performance. For robotic arms, the lack of expressive datasets has led to learning-based approaches, such as autoencoders \cite{26} and GANs \cite{27}, to reconstruct robotic motions from human motion dataset. However, these techniques often rely on randomly generated or industrial datasets as base robotic motions, which may lack expressiveness and compromise the final results.

\subsection{Motion Optimization Methods}
To enable realistic robot interaction, expressive motions must be optimized for collision-free, dynamically feasible execution while preserving style. Methods like TrajOpt 
 \cite{31}, STOMP \cite{kk}, and CHOMP \cite{CHOMP} optimize paths through functional constraints but often lose motion style due to global optimization. Whole-Body Control (WBC) \cite{WBC} enables real-time local obstacle avoidance and style preservation but requires high-performance hardware and complex modeling, limiting its use in low-cost robots. In this work, we propose a novel method that enables real-time obstacle avoidance on low-cost robots while effectively preserving trajectory shape and motion style.

To achieve real-time and generalizable expressive motion generation for 6-DOF companion robotic arms in real-world environments, Our approach addresses these challenges by explicitly mapping human dance data to robotic motions, leveraging diffusion models for generalization, simplifying the network for real-time performance and optimizing motions with style preservation for feasibility.

\section{System Overview}
The motion generation framework RMG is illustrated in Fig. \ref{fig:1}. Our key insight is to use dance motions to construct the robotic dataset, as dance embodies well-designed, widely accepted expressive movements. The framework consists of three modules: (1) the Motion Transforming module, which maps human dance motions to robotic arm motions; (2) the Motion Diffusion Generation module, which learns to generate motions based on start, end, and duration conditions; and (3) the Motion Optimization module, which post-processes motions from the preceding modules to ensure smoothness, feasibility, and collision avoidance. We implemented this framework on the Mycobot 280, a low-cost (\$1098) UR desktop robotic arm with a maximum reach of 280 mm, making it ideal for human-robot interaction scenarios.

\section{Motion Transforming}\label{sec:trans}
To construct the expressive motions dataset for robotic arm , we utilized the open-source human dance dataset, Popdancet \cite{23}, which comprises over 12,819 seconds of high-quality human dance motion data. The dataset is formatted in the SPML human body model in 60 FPS.
\subsection{Transforming Process} 
The process, illustrated in Fig. \ref{fig:2}, maps human motion to robotic arm motion by focusing on expressive aspects such as regularity, fluidity, and interactivity. Given the robotic arm's limited degrees of freedom, we prioritize replicating the most interactive trajectory. The hand, with its high motion density and expressive significance, is central to this process. By concentrating on the wrist trajectory (the root of the hand), we simplify the motion while preserving its essential features. We then analyze the total velocities of the trajectories of Points 20 and 21, selecting the one with the higher magnitude as $t_{wrist}$, as it best captures the dynamic and interactive nature of the motion.

To map $t_{wrist}$, we select Point C on the robotic arm. This point is controlled by the first three joints (J1, J2, J3) and determines the arm’s primary posture.The base of the robotic arm (Point A) is aligned with Point 3 in the human system, ensuring the mapped trajectory $t_C$ covers most of the workspace while avoiding extreme spatial concentrations. $t_C$ is a scaled version of $t_{wrist}$, with Point 3 as the origin, ensuring solvability within the robotic arm's workspace.

Using inverse kinematics, we analytically solve $t_C$ to derive all feasible $\theta_1, \theta_2, \theta_3$ for J1, J2 and J3.
To resolve the multi-solution problem, we use the human elbow trajectory $t_{elbow}$
as a reference, mapping the human elbow (Point 18 or 19) to Point B on the robotic arm. We compute $t_B$ using forward kinematics and minimize the following function:

\begin{equation}
\begin{aligned}  
\theta_1^{*}, \theta_2^{*}, \theta_3^{*} = \arg \min_{\theta_1, \theta_2, \theta_3} \left( | {t_B} - {scale} \cdot {t_{\text{elbow}}} | \right)\\
\text{subject to} \quad{t_C} = {scale} \cdot {t_{\text{wrist}}}.
\label{eq:p }
\end{aligned}  
\end{equation}
This ensures that $t_B$ closely mimics $t_{elbow}$, promoting smooth and continuous motion. It also enhances expressivity by capturing the spatial relationships between the wrist, elbow, and Point 3.

To enhance interactivity, we use J4, J5, and J6 (which determine Point D's orientation) to mimic the orientation of the human head (Point 15). Additionally, we constrain Point D to face forward most of the time by controlling J4 and J5: 
\begin{equation}
\theta_4 = -(\theta_2 + \theta_3), \quad \theta_5 = -\theta_1
\label{eq:sss}
\end{equation}

Due to morphological differences, the mapped angles may cause self-interference. Motion optimization (Sec. \ref{sec:opt}) refines the mapped trajectories to ensure they are collision-free and smooth. This process generates a dataset of over 10,000 collision-free, compliant, and highly expressive motion trajectories for the robotic arm.

\begin{figure*}[t]
    \centering
    \includegraphics[width=0.9\textwidth]{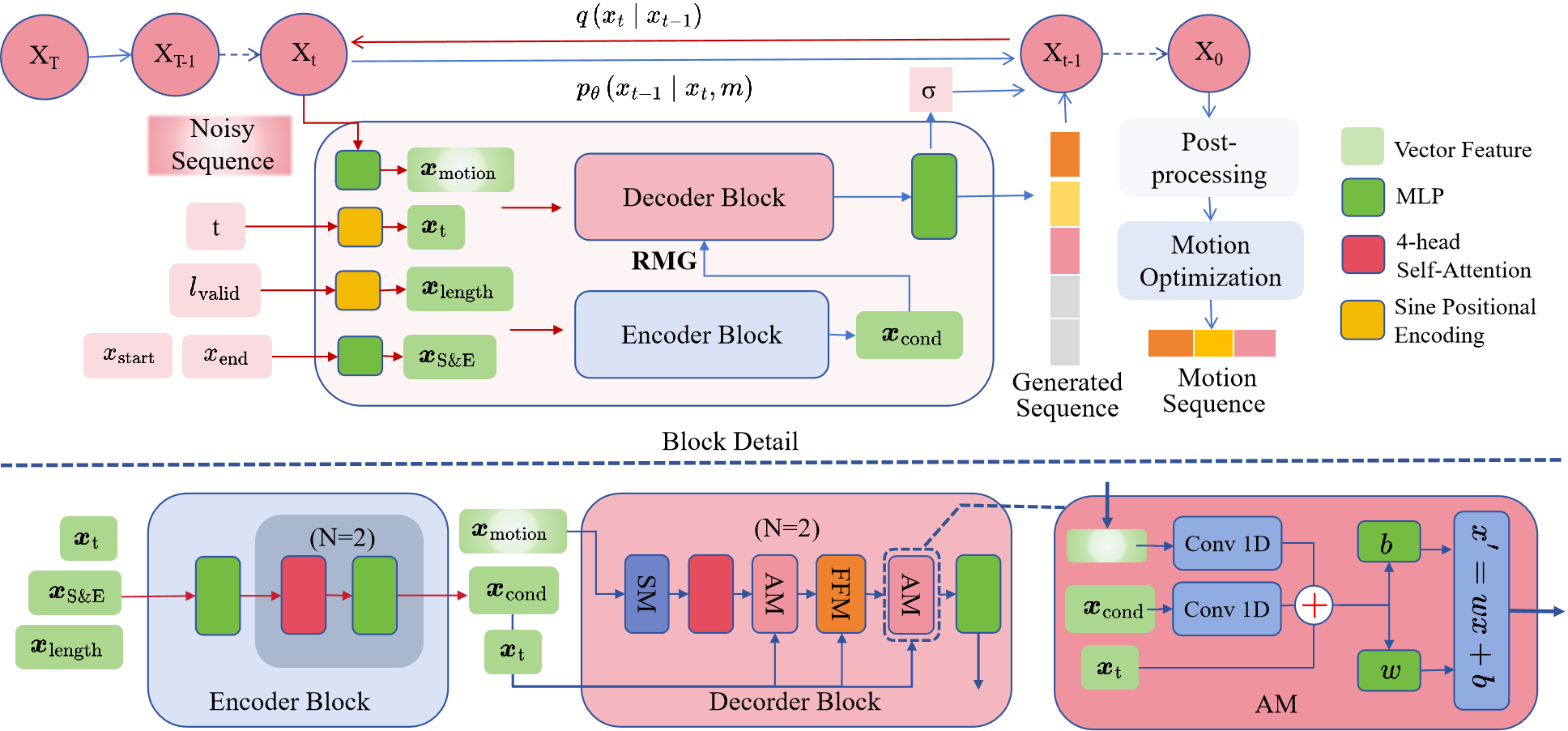}
    \caption{\textbf{Motion Diffusion Generation Overview.} The J Model and C Model share the same model structure. This system learns to denoise motion sequences from time \( t = T \) to \( t = 0 \), where \( t \) represents the diffusion steps. $l_{valid}$ is the valid sequence length to determine the motion time. In the C Model, start state \( x_{\text{start}} \) and end state \( x_{\text{end}} \) denote coordinate values, while in the J Model, they correspond to $\theta_1$, $\theta_2$, $\theta_3$. All input conditions are converted into vector feature \( x \in \mathbb{R}^{b \times h} \), and the noise sequence is converted to \( x_{\text{motion}} \in \mathbb{R}^{b \times l \times h} \), where \( L\) is the whole sequence length and \( h \) is set to 256 as the hidden layer dimension. \( N \) refers to the stack number. The postprocessing of the denoised sequence can be found in Sec. \ref{sec:processing}.}
    \label{fig:3}
\end{figure*}

\section{Motion Diffusion Generation}\label{sec:diff}
Given the robotic motion dataset from motion transforming (Sec. \ref{sec:trans}), we train two diffusion models: the J Model for motion generation in joint space and the C Model for Cartesian space. The pipeline is shown in Fig. \ref{fig:3}. Given an arbitrary start state $x_{start}$ , end state $x_{end}$, and motion duration (represented by the $l_{valid}$ ), these models synthesize smooth and expressive motion trajectories.

\subsection{Data Preprocessing and Postprocessing}

\paragraph{Preprocessing} We preprocess motion data from our robotic arm dataset, initially recorded in joint space. The raw data is segmented into motion sequences of 120 to 400 frames based on zero-velocity points. For real-time applicability, we downsample the data from 60 Hz to 12 Hz and pad all trajectories to 80 time steps using a masking strategy. In the C Model, positions of $t_C$ is extracted as input, which contain the message of $\theta_1$, $\theta_2$, $\theta_3$. By focusing exclusively on Point C, we streamline the $x_{\text{start}}$  and $x_{\text{end}}$ using only Point C’s position and enhance computational efficiency. In the J Model, $\theta_1$, $\theta_2$, $\theta_3$ are directly used as input features, which also inherently define $t_C$. To enhance temporal consistency and smoothness, we compute velocity components via finite differences and concatenate them with the original data, yielding the following structured representations:\newline
Joint Space Representation:
\begin{equation*}
X_{\text{angle}} \in \mathbb{R}^{l \times 6}: [\theta_1, \theta_2, \theta_3, V_{\theta_1}, V_{\theta_2}, V_{\theta_3}].
\end{equation*}
Cartesian Space Representation:
\begin{equation*}
X_{\text{cartesian}} \in \mathbb{R}^{l \times 6}: [X, Y, Z, V_x, V_y, V_z].
\end{equation*}
In this work, $l$ refers to 80.

\paragraph{Postprocessing} \label{sec:processing}Given the denosied sequence, we first extract the valid positional segment \( X' \in \mathbb{R}^{l_{\text{valid}} \times 3} \). The start and end points are then precisely aligned through interpolation to ensure trajectory continuity. For the C Model, joint angles are computed using inverse kinematics. To simplify computation and maximize interactivity, angles for J4, J5, and J6 are configured to maintain the end effector's forward-facing orientation. Finally, the angular motions \( X \in \mathbb{R}^{l_{\text{valid}} \times 6} \) are optimized using the algorithm described in Sec. \ref{sec:opt} to ensure smoothness and collision avoidance.

\subsection{Diffusion Framework}
Diffusion models are widely used for high-quality motion generation due to their ability to capture spatiotemporal relationships in long sequences. In this study, we employ Denoising Diffusion Implicit Models (DDIM) \cite{35}, which improves inference efficiency by reducing the number of required time steps through implicit sampling. However, DDIM's limited consideration of variance can restrict sample diversity. To address this, we integrate the improved Denoising Diffusion Probabilistic Models (iDDPM) method \cite{36} during training, which incorporates a variational lower bound loss (vlb). The vlb loss minimizes the discrepancy between the generated and true data distributions:
\begin{equation}   
\begin{aligned}  
L_{vlb}(t) &= \mathbb{E}_{x_t \sim q} \left[ -\log p(x_t | x_{t-1}) \right] \\
&\quad + D_{KL}(q(x_{t-1} | x_t) \| p(x_{t-1} | x_t))  
\end{aligned}  
\label{eq:vlb}
\end{equation}
The first term is the negative log-likelihood, which evaluates the model's prediction accuracy, while the second term is the Kullback-Leibler divergence, ensuring the generated distribution aligns closely with the true data distribution. By optimizing both mean and variance, this approach enables the generation of more diverse and high-quality samples.





\subsection{Encorder block}
The encoder block is designed to integrate conditional features. Initially, we sequentially concatenate the conditional vectors and process them through MLP to extract and combine essential features from these vectors. Following this, we apply a 4-head self-attention layer to capture the intricate relationships between the extracted features, thereby enhancing the model's expressive capability.

\subsection{Decoder Block}
The decoder block is responsible for decoding action sequences with style characteristics under specific conditions, utilizing various modules.

\subsubsection{Spatial Module (SM)} 
SM is designed to capture spatial relationships between different directions or joints using a 1D convolution. The input vector  \( x_{\text{motion}} \in \mathbb{R}^{b \times l \times h} \) is first transposed before being passed into the SM. This transformation generates an attention map that focuses on the spatial aspects of the data.

\subsubsection{Self-attention}
To capture temporal relationships within the input sequence, a 4-head self-attention layers is employed. The positionally encoded \( x_{\text{motion}} \) serves as both the query and key, while the original 
\( x_{\text{motion}} \)  acts as the value. These components are processed through the classical attention mechanism \cite{37}.


\subsubsection{Alignment Module (AM)} 
We adopted the AM in POPDG \cite{23}, as illustrated in Fig. \ref{fig:3}. It integrates motion features with conditions via 1D convolution and linear transformation, enhancing feature-condition alignment and improving the accuracy and coherence of generated sequences.

\begin{figure}[t]
    \centering
    \includegraphics[width=0.75\linewidth,keepaspectratio]{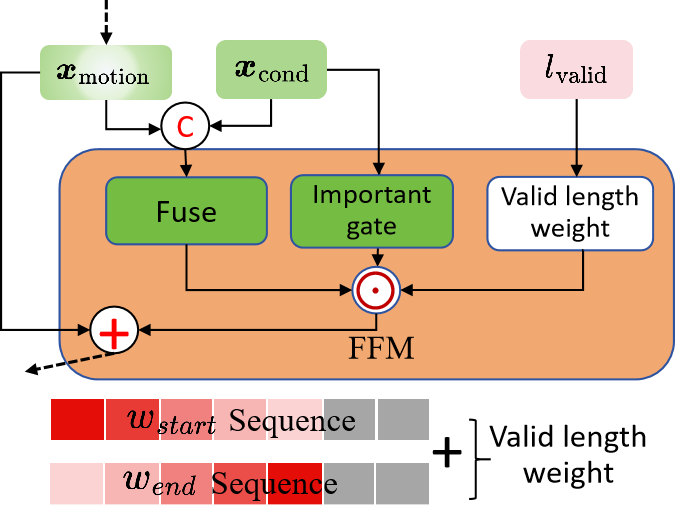}
    \caption{\textbf{Feature Fusion Module.} In the 
\( w \) sequence, the gray areas indicate invalid parts, and the color intensity in the valid sections reflects the weight. The sum of the two sequences gives the valid length weight.}
    \label{fig:4}
\end{figure}

\subsubsection{Feature Fusion Module (FFM)}
Shown in Fig. \ref{fig:4}, FFM deeply integrates action features with conditional features. To ensure a smooth transition in the generated motion sequence from a start point to an endpoint with minimal deviation, the module places special emphasis on the sequence points near these critical points during the fusion process. It utilizes a dynamic weighting mechanism that adjusts the fusion of features based on $l_{valid}$.  Specifically, the module calculates the sorting distance from each sequence point to both the start and end of the valid sequence, and then applies a Gaussian function to generate the corresponding weights:  
\begin{equation}  
\begin{array}{rl}  
w_{\text{end}} & = \exp\left( -\frac{2 \cdot \text{end\_distance}^2}{\sigma^2} \right) \\
w_{\text{start}} & = \exp\left( -\frac{2 \cdot \text{start\_distance}^2}{\sigma^2} \right)  
\end{array}  
\end{equation}  
where \(\sigma\) is a parameter related to the valid sequence length. The closer a position is to either the start or end point, the higher the corresponding weight will be. The final valid length weight is the sum of \(w_{\text{start}}\) and \(w_{\text{end}}\). 

Furthermore, the importance gate processes the conditional features to enhance focus on key attributes. Subsequently, the condition and sequence features are fused through a transformation network. The final output combines the original sequence features with the weighted fused features, thereby enhancing the model’s adaptability, expressiveness, and comprehension of complex temporal data, while also providing improved interpretability.

\subsection{Loss Function }
During training, additional losses complement the VLB loss (Eq. \ref{eq:vlb}). Let \( \boldsymbol{x} \)  and \( \hat{\boldsymbol{x}}_{\theta, t} \) denote the original and generated data at diffusion step \( t \), with \([ \text{pad} ]\) and  \([ \text{valid} ]\) representing masked and valid portions, respectively.

Loss for the masked portion \( \mathcal{L}_{\text{pad}} \):
\begin{equation}
\mathcal{L}_{\text{pad}} = \mathbb{E}_{\boldsymbol{x}, t} \left[ \left\| \boldsymbol{x}[\text{pad}] - \hat{\boldsymbol{x}}_{\theta,t}[\text{pad}] \right\|_2^2 \right]
\end{equation}

Loss for the valid portion \( \mathcal{L}_{\text{pv}} \): where the first three dimensions represent position loss and the last three dimensions represent velocity loss:
\begin{equation}
\mathcal{L}_{\text{pv}} = \mathbb{E}_{\boldsymbol{x}, t} \left[ \left\| \boldsymbol{x}[\text{valid}] - \hat{\boldsymbol{x}}_{\theta,t}[\text{valid}] \right\|_2^2 \right]
\end{equation}

By differentiating the valid sequence portion, we obtain velocity and acceleration losses \( \mathcal{L}_{\text{va}} \). Unlike the velocity loss in \( \mathcal{L}_{\text{pv}} \), this velocity loss focuses on the smoothness and continuity of the position part:
\begin{equation}
\mathcal{L}_{\text{va}} = \mathbb{E}_{\boldsymbol{x}, t} \left[ \left\| \boldsymbol{x'}[\text{valid}] - {\hat{\boldsymbol{x}}_{\theta,t}'}[\text{valid}] \right\|_2^2 \right]
\end{equation}

The start and end errors of the valid sequence portion are treated as start and end losses \( \mathcal{L}_{\text{start}} \) and \( \mathcal{L}_{\text{end}}\):
\begin{gather}
\mathcal{L}_{\text{start}} = \mathbb{E}_{\boldsymbol{x}, t} \left[ \left\| \boldsymbol{x}[\text{valid}][0] - \hat{\boldsymbol{x}}_{\theta,t}[\text{valid}][0] \right\|_2^2 \right]
\\
\mathcal{L}_{\text{end}} = \mathbb{E}_{\boldsymbol{x}, t} \left[ \left\| \boldsymbol{x}[\text{valid}][-1] - \hat{\boldsymbol{x}}_{\theta,t}[\text{valid}][-1] \right\|_2^2 \right]
\end{gather}

The final total loss is expressed as:
\begin{equation} 
\mathcal{L}_{\text{total}} = w_0\mathcal{L}_{\text{vlb}}+w_1 \mathcal{L}_{\text{pv}} + w_2 \mathcal{L}_{\text{va}} + w_3 \mathcal{L}_{\text{pad}} + w_4 \mathcal{L}_{\text{start}}+ w_5 \mathcal{L}_{\text{end}}
\end{equation}
$w$ denotes the weights of the losses.

\section{Motion Optimization} \label{sec:opt}
To ensure that the robotic arm's motions are collision-free, smooth while preserving the motion style. we optimized the motion trajectories generated by mapping algorithms and diffusion models.

We developed an obstacle avoidance algorithm using Particle Swarm Optimization (PSO) \cite{33}, integrated with Curobo \cite{32} for efficient parallel collision detection in joint space. The algorithm evaluates collision costs  $Curobo(x_n)$ for each point $x_n$ in a motion sequence \( X \in \mathbb{R}^{N \times 6} \)  to identify collision points.For each of these points, we update 50 particles and compute their fitness $f$:
\begin{align}
f = 10000 \times  Curobo(x_n)  + \Delta \theta\label{eq:fitness}
\end{align}
The fitness function is designed to strictly avoid collisions while minimizing changes in joint angles to preserve motion style. Particle positions are updated iteratively using the PSO algorithm until the fitness values of all collision points fall below a predefined threshold. The specific algorithm implementation can be summarized by Algorithm \ref{alg:pso_obstacle_avoidance}.

\begin{algorithm}
\caption{PSO for Collision Avoidance}
\label{alg:pso_obstacle_avoidance}
\begin{algorithmic}[1]
\State \textbf{Input:} $X \in \mathbb{R}^{N \times 6}, w, c_1, c_2$
\State $r_1, r_2 \sim \text{Uniform}(0, 1),f_{gbest} = \infty$
\For{each $n \in C = \{ n \mid Curobo(x_n) > 0 \}$}
    \State Initialize particles: $p_i, v_i, pbest_i = p_i, f_{pbest_i} = \infty$
    \Repeat
        \For{$i = 1$ \textbf{to} $n$}  
            \State $f_i = 10000 \times Curobo(p_i) + \Delta \theta_i$
            \If{$f_i < f_{pbest_i}$}
                \State $pbest_i = p_i, f_{pbest_i} = f_i$
            \EndIf
            \If{$f_i < f_{gbest}$}
                \State $gbest = p_i, f_{gbest} = f_i$
            \EndIf
            \State 
            $v_i = w \cdot v_i + c_1 \cdot r_1 \cdot (pbest_i - p_i)$
            \State $ \quad \quad + c_2 \cdot r_2 \cdot (gbest - p_i)$
            \State $p_i = p_i + v_i$
        \EndFor
    \Until{$f_{gbest} < \text{threshold}$}
\EndFor
\end{algorithmic}
\end{algorithm}
If the fitness values remain above the threshold and collisions persist after 30 iterations, problematic points are removed. We then apply B-spline fitting and Gaussian filtering to smooth the trajectory, followed by a recheck for collisions. 
Finally, we use the TOPP-RA algorithm \cite{34} from MoveIt to replan the motion timing, avoiding sudden accelerations or decelerations and ensuring smooth execution within the robotic arm’s velocity and acceleration limits.

\section{Experiments and Analysis}
In this section, we conduct experiments and analyze each module separately. All training and computations are performed on a computer with an Nvidia 4070 GPU, with results demonstrated in the Gazebo simulation environment.

\subsection{Motion Optimization Effects} \label{sec:fid}
We compare our motion optimization algorithm (Sec. \ref{sec:opt}) with Curobo's TrajOpt in terms of speed and motion style preservation. Motion style preservation is evaluated by comparing the similarity between the optimized robotic trajectories and the original human trajectories. Specifically, we compare the robot's $t_C$ with the human's scaled $t_{wrist}$ and the robot's $t_B$ with human's scaled $t_{elbow}$, as these represent the primary motions. Comparing with the original human trajectories provides a unified benchmark for subsequent experiments. 

Motion features are assessed using a vector \( X \in \mathbb{R}^{35} \), derived from frequency characteristics, motion features, and trajectory morphology. Similarity in motion features is measured using Frechet Inception Distance (FID) \cite{29} for distribution distance, while trajectory sequence similarity is evaluated using average Dynamic Time Warping (DTW) \cite{30}. DTW finds the optimal matching path between two trajectories and calculates the minimum cumulative distance after normalization and length effect elimination.

Results in Table \ref{lab1} show that, unlike TrajOpt—which optimizes the entire trajectory simultaneously, altering previously collision-free points and degrading path shape and style—our approach better preserves trajectory style, as evidenced by significantly lower FID and DTW scores. Furthermore, leveraging GPU parallel computing, our method achieves a 300× speedup over serial computation and is over two times faster than TrajOpt, fully meeting real-time requirements.


\begin{table}[t]
    \centering
    \begin{tabular}{l|ccccc}
    \toprule
          Method&
          $\text{FID}_\text{C}\downarrow$& $\text{FID}_\text{B}\downarrow$ & $\text{DTW}_\text{C}\downarrow$ & $\text{DTW}_\text{B}\downarrow$ & 
          $T_{opt}\downarrow$\\ \midrule
        TrajOpt \cite{31}& 159.97 & 473.34 & 0.38 & 1.31 & 0.65\,s  \\
        RMG(ours) & \textbf{6.17} & \textbf{94.72} & \textbf{0.11} & \textbf{0.75} & \textbf{0.29\,s}  \\ \bottomrule

    \end{tabular}
    \caption{\textbf{Algorithm Comparison.}  Subscript C denotes the robot's $t_C$ metrics and subscript B denotes robot's $t_B$ metrics.  TrajOpt \cite{31} is an algorithm in Curobo \cite{32} and tends to prioritize minimum-time collision-free paths, resulting in a significant loss of human stylistic elements. $T_{opt}$ represents the optimization time for trajectories with an average of 20\% collision points and a sequence length of 200. Additionly, RMG's average optimization time is 105 seconds without parallel processing. } 
    \label{lab1}
\end{table}
\subsection{Motion Transforming Effects}
The quality of motion transforming in Sec. \ref{sec:trans} depends on motion expressivity and mapping similarity.
\subsubsection{Similarity} 
Table \ref{lab1} also reflects the similarity between the final robotic transformed motions and the original human motions, as $t_C$ vs. $t_{wrist}$ and $t_B$ vs. $t_{elbow}$ represent the primary mapping relationships. Our method achieves significantly lower FID and DTW values compared to the TrajOpt group, which optimizes the same mapped motions and serves as a baseline for non-similarity. This demonstrates its ability to preserve dance characteristics. The FID and DTW values for $t_{C}$ are notably lower than those for $t_{B}$, indicating that $t_{C}$ (mapped from the wrist) closely resembles the original motion. In contrast, $t_{B}$ captures only partial elbow characteristics because $t_{elbow}$ serves as an auxiliary reference and is not always reachable due to structural constraints, which explains its higher FID and DTW values.

\subsubsection{Expressivity}\label{sec:dist}
We evaluate motion expressivity by comparing the our method with RoboGroove \cite{bib4}, which uses torso joint angles (Points 0, 6, 15) as robotic joint angles. Motion diversity is a key indicator of expressivity. To quantify this, we compute the average Euclidean distance between 40 mapped robotic motions in the feature space of $t_B$ and $t_C$ (the same as used for FID in Sec. \ref{sec:fid}).

Smoothness is another key characteristic, measured using the average Curvature Change Rate (CCR) of $t_C$. CCR quantifies the rate of curvature variation along a trajectory, where a lower value indicates smoother motion. It is calculated as:
\begin{equation}  
    \text{CCR} = \frac{1}{N} \sum_{i=1}^{N} \left| \frac{d^2 \mathbf{r}(s_i)}{ds^2} \right|
\end{equation}  
where \( \mathbf{r}(s_i) \) is the position at arc length \( s_i \), and \( N \) is the number of sampled points. 

Other aspects of expressivity are assessed through a user study with 20 participants. Participants rate the generated motions form both methods on biologicality (lifelike and fluid motions), playfulness (dynamic, graceful, and varied motions), and interactivity (engagement with a hypothetical person) on a scale of 1 to 5. 

Results in Table 2 show that RMG outperforms RoboGroove in all categories except CCR, where RMG's score is slightly higher. However, this is acceptable because RMG achieves a higher Dist score, indicating increased motion complexity. This demonstrates that hand-based mapping yields better expressivity than torso-based mapping, while also achieving a better balance between motion complexity and smoothness.

\begin{table}[t]
    \centering
        {
    \begin{tabular}{c|ccccc}
    \toprule
        \multirow{2}{*}{Method}&
        \multirow{2}{*}{Dist$\uparrow$}&
        \multirow{2}{*}{\makecell{CCR $\downarrow$\\ $[\text{m}^{\scriptsize{-2}}]$}}&
        \multirow{2}{*}{\makecell{Bio- \\ logicality $\uparrow$} } & 
        \multirow{2}{*}{\makecell {Playful- \\ness $\uparrow$} }& 
        \multirow{2}{*}{\makecell{Inter-\\activity $\uparrow$ }}
          \\\\ \midrule
        \multirow{2}{*}{\makecell{Robo-\\Groove\cite{bib4}}} &
        \multirow{2}{*}{5.52} & 
        \multirow{2}{*}{\textbf{80.02}} & 
        \multirow{2}{*}{3.2} & 
        \multirow{2}{*}{2.6} & 
        \multirow{2}{*}{2.75}  \\ \\
        \multirow{2}{*}{RMG(ours)} &
        \multirow{2}{*}{\textbf{8.06}}& 
        \multirow{2}{*}{91.63} &
        \multirow{2}{*}{\textbf{4.2}} &
        \multirow{2}{*}{\textbf{4.25}} & 
        \multirow{2}{*}{\textbf{4.3}}  \\\\ \bottomrule
    \end{tabular}}
        \caption{\textbf{Expressiveness Evaluation For Motion Transforming.} } 
    \label{lab2}
\end{table}

\subsection{Motion Diffusion Generation Effects}
In motion diffusion generation, both the J Model and C Model were trained for 2k epochs, including a 20-epoch warm-up phase. Optimization used the Adan optimizer with a learning rate of 2e-4 and a weight decay rate of 1.2e-4.

Since our model focuses on generating expressive motion trajectories rather than traditional path planning tasks aimed at finding optimal paths, we evaluate overall performance in both planning ability and expressivity. Similar works include ELEGNT \cite{bib2} and Osorio et al.'s work \cite{26}, but neither is open-source, and their tasks are different from ours. We only compare and analyze the J model and C model by generating robot motions for 40 test set conditions, using the corresponding robotic dataset motions obtained from motion transforming as the oracle. The results are presented in TABLE \ref{tab3}.

\begin{figure*}[htbp]
\centering
\begin{minipage}[t]{0.48\textwidth}
\centering
\includegraphics[width=0.98\textwidth]{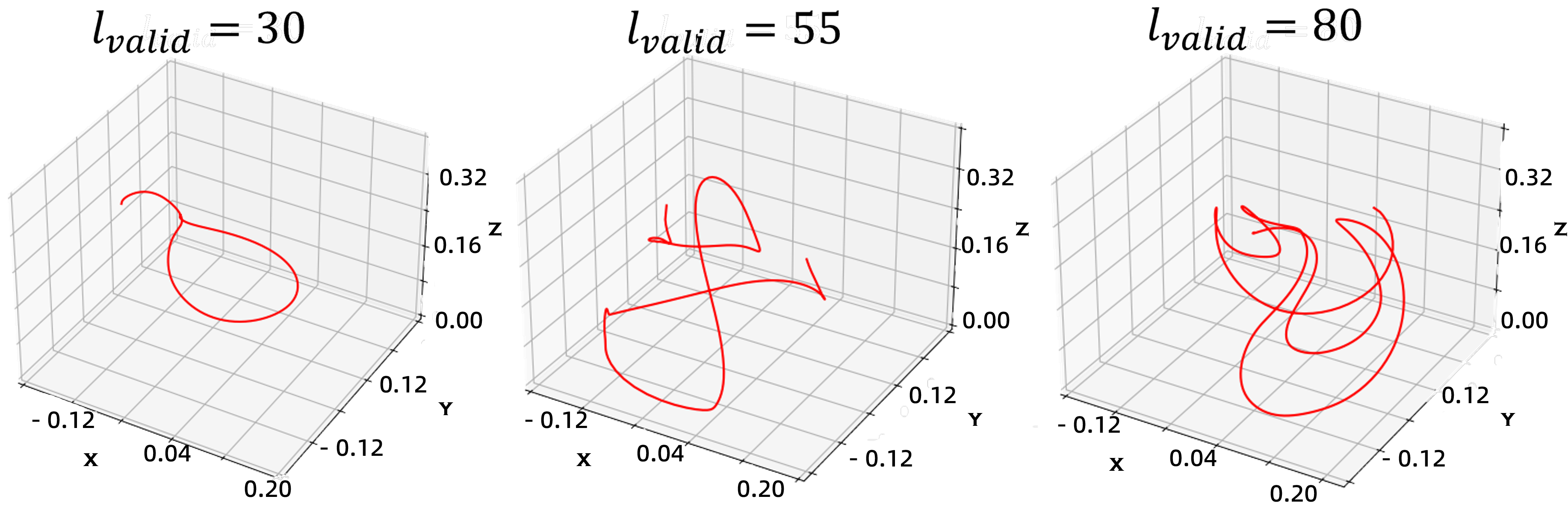}
\caption{\textbf{Point C trjectory $t_C$ in the J Model}  }
\label{5}
\end{minipage}
\begin{minipage}[t]{0.48\textwidth}
\centering
\includegraphics[width=0.98\textwidth]{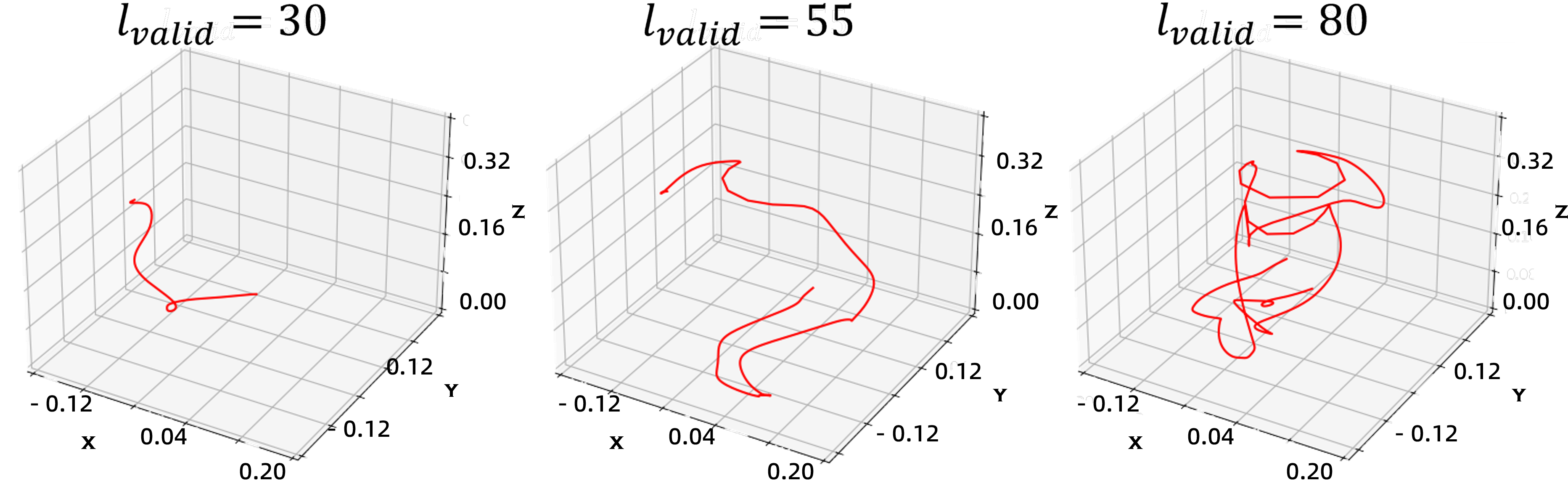}
\caption{\textbf{Point C trjectory $t_C$ in the C Model}}
\label{6}
\end{minipage}
\vspace{1.0em}
\end{figure*}

\begin{table*}[t]
    \centering
    \begin{tabular}{lccccccccc}
    \toprule
            ~ & \multicolumn{2}{c}{\textbf{Motion Quality}} & \multicolumn{1}{c}{\textbf{Diversity}} & \multicolumn{1}{c}{\textbf{Smoothness}} & \multicolumn{5}{c}{ \textbf{Planning Ability}} \\ 
        \cmidrule(lr){2-3} \cmidrule(lr){4-4}  \cmidrule(lr){5-5}  \cmidrule(lr){6-10}
        ~ & 
        $\text{FID}_\text{C}\downarrow$ & $\text{FID}_\text{B}\downarrow$ & 
        $\text{Dist}\uparrow$ & CCR$[\text{m}^{\scriptsize{-2}}]$ &
        $|\Delta_\text{start}|\downarrow$ & $|\Delta_\text{end}|\downarrow$ & 
        $\eta_\text{opt} \downarrow$ & 
        $T_{gen}$ & 
        $|\Delta_{T_{Motion}}|\downarrow$  \\ 
        \midrule
        Robotic Dataset & 6.17 & 94.72 & 8.06 & 91.63 & - & - & - & - & -  \\ 
        J Model & \textbf{19.86} & \textbf{114.72} & 8.96 & \textbf{83.10} & 0.05\,rad & 0.22\,rad &\textbf{ 4.1\%} & \textbf{0.37\,s} & \textbf{0.33\,s} \\ 
        C Model & 81.75 & 287.16 & \textbf{13.32} & 127.8  & 8.71\,mm & 13.30\,mm & 17.9\% & 0.46\,s & 1.86\,s  \\ 
        \bottomrule
    \end{tabular}
    \caption{\textbf{Motion Generation Evaluation.} 
    The robitc dataset is generated from motion transformation (Sec. \ref{sec:trans}). Subscript C denotes the robot's $t_C$ metrics and subscript B denotes robot's $t_B$ metrics.
    \( \Delta_\text{start} \) and \( \Delta_\text{end} \) represent the average deviation from the given start and end points before fine tuning. \( \eta_\text{opt} \) indicates the proportion of points in the original sampled path that need adjustment. \( T_{\text{gen}} \) is the motion generation time. \( \Delta_{T_{\text{Motion}}} \) is the  motion time error  in simulation from the original motion.}
    \label{tab3}
\end{table*}

\subsubsection{Planning Ability}    
\paragraph{Errors in Condition} Both models achieve minimal errors in start and end points ( $\Delta_{\text{start}}$ and $\Delta_{\text{end}}$ ). Additionally, the motion duration deviation $\Delta T_{\text{Motion}}$ in the J Model is only 0.34 seconds from the dataset's reference motion time, 6 times smaller than the C Model's, demonstrating superior temporal control.
\paragraph{Optimization Rate} the raw motion sequences contain points that require optimization due to collisions and inverse kinematics failures, quantified by the optimization rate $\eta_{\text{opt}}$. The J Model achieves a 4.4× lower $\eta_{\text{opt}}$, indicating a superior understanding of the free configuration space. In contrast, the C Model relies more heavily on post-processing optimization, likely due to its training data being restricted to the Point C trajectory without global information. 
\paragraph{Speed}
Both models achieve real-time performance with $T_{gen}$ under 0.5s. However, the C Model requires 1.4 times longer due to additional kinematic processing.

Overall, the J Model outperforms the C Model in planning ability, demonstrating superior temporal control, lower optimization requirements, and faster motion generation.


\pagebreak
\subsubsection{Motion Expressivity}  
\paragraph{Motion Quality} 
We evaluated the feature distribution distance between human motions and generated robot motions using FID (Sec. \ref{sec:fid}) to assess the inheritance of human dance motion expressivity. The J Model's FID is slightly higher than the baseline, indicating improved generalization while preserving motion characteristics. In contrast, the C Model exhibits significantly higher FID values, especially for 
$\text{FID}_\text{B}$, likely due to its reliance on post-processing optimization and the lack of Point B data in training.

\paragraph{Diversity}
Diversity was assessed by computing Dist (Sec. \ref{sec:dist}). The J Model exhibits slightly higher Dist than the robotic dataset, reflecting its generalization capability. In contrast, the C Model's Dist value is substantially higher, correlating with its elevated FID. This increased diversity in the C Model results from trajectory optimization required for 17.9\% of points, leading to more chaotic trajectories.

\paragraph{Smoothness} The CCR of $t_C$ is used to evaluate the motion smoothness. The J Model's CCR closely matches the robotic dataset, demonstrating a better balance of smoothness and motion complexity. However, the C Model exhibits significantly higher CCR values, which may lead to less natural motion transitions.


\paragraph{Visualization and User Study}
We visualize motions generated under different $l_{valid}$ (30, 55, 80) by plotting $t_C$. As shown in Fig. \ref{5} and \ref{6}, motions become more complex as $l_{valid}$ increases, highlighting the impact of length conditions on motion generation. The J Model produces smoother and more diverse trajectories, while the C Model generates more rigid and less natural motions, aligning with the CCR values.

In the user study, 80\% of participants preferred the motions generated by the J Model, a preference consistent with the visual characteristics of the trajectories. 


In conclusion, while both models can generate expressive motion planning in real-time, the J Model, with its better understanding of the configuration space during training, produces higher-quality and more expressive motions.

\begin{table}[!ht]
    \centering
    \scriptsize 
    \setlength\tabcolsep{1.25pt}
    \scalebox{1.0}{
    \begin{tabular}{p{5em}|l|cccccc}
    \toprule
         Model &
         Method &
         $\text{FID}_\text{C}\downarrow$ & 
         $\text{FID}_\text{B}\downarrow$ & $|\Delta_\text{start}|\downarrow$ & $|\Delta_\text{end}|\downarrow$ & 
         $\eta_\text{opt} \downarrow$ &  
         $|\Delta_{T_{M}}|\downarrow$ \\ \midrule
        \multirow{4}{5em}{\textbf{J Model}} & RMG & 19.86 & 114.72 & 0.05\,rad & 0.22\,rad & 4.1\% & 0.33\,s  \\ 
         & w/o SM & 23.92 & 118.53 & 0.08\,rad & 0.14\,rad & 9.1\% & 0.37\,s  \\ 
         & w/o FFM & 22.72 & 115.92 & 0.09\,rad & 0.66\,rad & 7.5\% & 0.40\,s  \\
         & w/o AM & 28.32 & 124.98 & 0.24\,rad & 0.31\,rad & 12.1\% & 0.51\,s  \\ \midrule
        \multirow{4}{5em}{\textbf{C Model}} & RMG & 81.75 & 287.16 & 8.71\,mm & 13.3\,mm & 17.9\% & 1.86\,s  \\ 
         & w/o SM & 137.47 & 380.84 & 8.52\,mm & 15.1\,mm & 36.7\% & 2.26\,s  \\ 
         & w/o FFM & 90.85 & 300.14 & 10.1\,mm & 48.3\,mm & 26.6\% & 2.54\,s  \\
         & w/o AM & 198.98 & 458.26 & 7.93\,mm & 9.22\,mm & 39.7\% & 2.67\,s  \\ \bottomrule
    \end{tabular}}
    \caption{\textbf{Ablation Study.} Without SM, performance declines across all metrics, with $\eta_\text{opt}$ doubling. Without FFM, $\Delta_\text{end}$ increases by more than threefold. Without AM, all metrics deteriorate, though the J Model shows less degradation compared to the C Model.}
    \label{tab4}
\end{table}

\subsection{Ablation Study for Motion Diffusion Generation}
We conducted ablation experiments in our Generation Models to evaluate the effectiveness of the key design components: SM, AM, and FFM. As shown in TABLE \ref{tab4}, the J Model demonstrates greater robustness compared to the C Model. With SM, the model better captures the style and spatiotemporal features of trajectories, leading to improved motion quality (FID) and planning ability ($\Delta_\text{start}$, $\Delta_\text{end}$, $\eta_\text{opt}$,  $\Delta_{T_{M}}$). The FFM enhances the model's ability to fuse $x_{start}$, $x_{end}$ and $l_{valid}$, significantly reducing $\Delta_\text{start}$ and $\Delta_\text{end}$. Additionally, AM improves the alignment of condition and trajectory characteristics and achieve a better balance between motion quality and planning capability, resulting in overall superior performance.

\begin{figure}[!ht]
    \centering
    \includegraphics[width=0.8\linewidth, height=5cm]{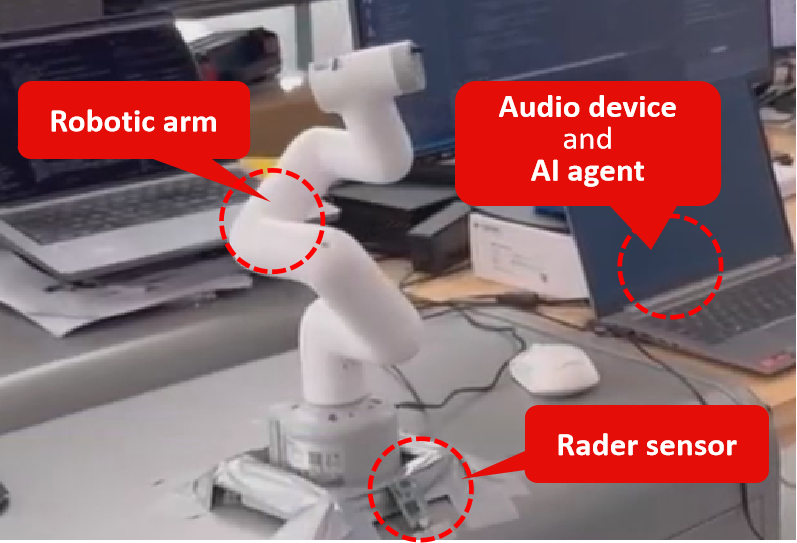}
    \caption{\textbf{Prototype Overview.} }
    \label{deploy}
\end{figure}

\section{Deployment}
We successfully deployed the generative model on the low-cost desktop robotic arm Mycobot280 for real-time interaction. As shown in Fig. \ref{deploy}, the arm is equipped with a radar sensor for gesture recognition, an AI agent, and audio devices for voice interaction. When the sensor detects a gesture, the arm generates an expressive motion. During AI interactions, the agent estimates speech duration and randomly selects an end state, which is fed into the generation model to create a corresponding expressive motion. With our method, the motion is non-repetitive and can be generated in real-time from any start and end point. The results demonstrate that interactions are engaging, particularly during motion demonstrations accompanied by voice.


\section{Conclusion}
In this work, we demonstrated real-time expressive motion interaction using a 6-DOF companion robotic arm. We introduced a mapping method to extract expressive features from human motion and created a corresponding robotic arm expressive motion dataset. Additionally, we developed a motion optimization algorithm to ensure smoothness, avoid self-collisions, and preserve motion style. Furthermore, we designed a iDDPM-based network capable of generating expressive motions for the robotic arm based on a start state, end state, and motion duration. We trained both Cartesian space and joint space models, with both meeting real-time interaction requirements. With global pose information, the joint space model outperformed the Cartesian space model due to its superior understanding of the configuration space.

There are some limitations in our study. In the prototype, due to the robot's limited performance and challenges in maintaining precise speed control, it frequently overshoots, stutters, and struggles to track trajectories accurately. Additionally, motion generation is constrained by predefined trajectory endpoints and durations, lacking emotional classification, which limits its ability to produce contextually appropriate responses. Future work will focus on enabling the robotic arm to autonomously select emotional responses, determine motion endpoints, and adjust motion duration based on contextual analysis, paving the way for more natural and intelligent human-robot interactions.

\addtolength{\textheight}{-12cm}   








\footnotesize
\bibliographystyle{IEEEtran}
\bibliography{reference}
\end{document}